\documentclass{llncs}

\usepackage{listings}
\usepackage{todonotes} 
\usepackage{color}
\usepackage{courier}
\usepackage{booktabs}
\usepackage{wrapfig}
\usepackage{hyperref}
\usepackage[utf8]{inputenc}
\usepackage{xcolor}

\colorlet{punct}{red!60!black}
\definecolor{delim}{RGB}{20,105,176}
\colorlet{numb}{magenta!60!black}

\hypersetup{
    pdfinfo={
        Title={Geospatial Reasoning with Shapefiles for Supporting Policy Decisions},
        Author={Henrique Santos, James P. McCusker, Deborah L. McGuinness},
        Keywords={shapefiles, reasoning, radio spectrum policies}
    }
}

\definecolor{lightBlue}{RGB}{0, 153, 255}
\definecolor{lightRed}{RGB}{204, 0, 255}

\makeatletter
\newcommand*{\centerfloat}{
  \parindent \z@
  \leftskip \z@ \@plus 1fil \@minus \textwidth
  \rightskip\leftskip
  \parfillskip \z@skip}
\makeatother

\makeatletter
\DeclareRobustCommand\ttfamily
        {\not@math@alphabet\ttfamily\mathtt
         \fontfamily\ttdefault\selectfont\hyphenchar\font=-1\relax}
\makeatother
\DeclareTextFontCommand{\mytexttt}{\ttfamily\hyphenchar\font=45\relax}

\makeatletter
\renewcommand*{\@fnsymbol}[1]{\ensuremath{\ifcase#1\or *\or \dagger\or \ddagger\or
    \mathsection\or \mathparagraph\or \|\or **\or \dagger\dagger
    \or \ddagger\ddagger \else\@ctrerr\fi}}
\makeatother

\lstdefinelanguage{Manchester}
{
    sensitive = true,
    keywords = [1]{Class, EquivalentTo, SubClassOf},
    morekeywords = [2]{and, or},
    morekeywords = [3]{some, value},
    keywordstyle=[2]\textbf,
    keywordstyle=[2]\color{lightBlue}\textbf,
    keywordstyle=[3]\color{lightRed}\textbf,
    morestring=[b]"
}

\lstdefinelanguage{json}
{
    basicstyle=\normalfont\ttfamily,
    numbers=left,
    numberstyle=\scriptsize,
    stepnumber=1,
    numbersep=8pt,
    showstringspaces=false,
    breaklines=true,
    frame=lines,
    literate=
     *{0}{{{\color{numb}0}}}{1}
      {1}{{{\color{numb}1}}}{1}
      {2}{{{\color{numb}2}}}{1}
      {3}{{{\color{numb}3}}}{1}
      {4}{{{\color{numb}4}}}{1}
      {5}{{{\color{numb}5}}}{1}
      {6}{{{\color{numb}6}}}{1}
      {7}{{{\color{numb}7}}}{1}
      {8}{{{\color{numb}8}}}{1}
      {9}{{{\color{numb}9}}}{1}
      {:}{{{\color{punct}{:}}}}{1}
      {,}{{{\color{punct}{,}}}}{1}
      {\{}{{{\color{delim}{\{}}}}{1}
      {\}}{{{\color{delim}{\}}}}}{1}
      {[}{{{\color{delim}{[}}}}{1}
      {]}{{{\color{delim}{]}}}}{1},
}

\begin{document}

\renewcommand{\thelstlisting}{\arabic{lstlisting}}

\title{Geospatial Reasoning with Shapefiles for Supporting Policy Decisions}

\author{Henrique Santos
\and James P. McCusker
\and Deborah L. McGuinness
}


\institute{Rensselaer Polytechnic Institute, Troy NY, USA 12180} %

\maketitle
\begin{abstract}
Policies are authoritative assets that are present in multiple domains to support decision-making. They describe what actions are allowed or recommended when domain entities and their attributes satisfy certain criteria. It is common to find policies that contain geographical rules, including distance and containment relationships among named locations. These locations' polygons can often be found encoded in geospatial datasets. We present an approach to transform data from geospatial datasets into Linked Data using the OWL, PROV-O, and GeoSPARQL standards, and to leverage this representation to support automated ontology-based policy decisions. We applied our approach to location-sensitive radio spectrum policies to identify relationships between radio transmitters coordinates and policy-regulated regions in Census.gov datasets. Using a policy evaluation pipeline that mixes OWL reasoning and GeoSPARQL, our approach implements the relevant geospatial relationships, according to a set of requirements elicited by radio spectrum domain experts.
\end{abstract}


\section{Introduction}

Policies are commonly defined as decision-making assets that express one or more actions allowed or recommended under certain conditions. In the radio communications domain, policies are created to help manage the use of a limited electromagnetic spectrum. Many policies are location-specific, meaning that they are only applicable when the usage of the radio spectrum is to occur in specific geographic locations, as dictated by the policy. In the United States, the \textit{National Telecommunications and Information Administration Manual of Regulations and Procedures for Federal Radio Frequency Management}\footnote{\url{http://bit.ly/NTIA_Redbook}} (\textit{NTIA Redbook}) is a compilation of regulatory policies that define the conditions organizations, systems, and devices must satisfy to compatibly share radio spectrum while minimizing interference. Because policies in the NTIA Redbook regulate both commercial and federal spectrum usage, it is common to find military facilities, as well as regulations covering domestic and international locations.

The US Census Bureau publishes geospatial datasets about the United States, its territories, and points of interest, in its Census.gov data portal. The datasets contain high-definition polygons, usually in the \texttt{shapefile}~\cite{noauthor_esri_1998} format, of many locations referred by radio spectrum policies. Although this format is a popular choice for encoding geographical entities, its main use-case is to support data interchange among geographic information systems (GIS). The \texttt{shapefile} format usually requires the use of a GIS to allow operations over the data, including calculations and queries. Therefore, it is not very suitable for integrating with ontology-based applications.

We present an approach to allow ontology-based applications to leverage geospatial data in formats not easily accessible or referred from within ontology constructs, and to use this data to perform geospatial calculations. We implemented the approach to support automated radio spectrum policy decisions. This is accomplished by representing Census.gov relevant polygons in the GeoSPARQL~\cite{battle_enabling_2012} vocabulary, and by defining OWL~\cite{grau_owl_2008} classes that encode the policies' location rules. A policy evaluation pipeline that mixes OWL reasoning and GeoSPARQL leverages this model to elicit spatial relationships, providing high-definition spatial calculations. This approach was evaluated to perform well in terms of coverage of geospatial requirements, as elicited by domain experts. It is implemented as part of the Dynamic Spectrum Access (DSA) Policy Framework~\cite{santos_dsa_2020}, which was developed to serve as a machine-readable policy repository to support increased automation of policy evaluations.

\section{Transforming Census.gov \mytexttt{shapefiles}}

The majority of location-sensitive policies in the NTIA Redbook refer to these types of locations: military facilities, States, or Country. Because the policies are originally authored in natural language, and targeted at spectrum managers, they refer to locations by their names (e.g. ``Fairbanks'', ``Camp Parks''), without a comprehensive definition of the boundaries of such regions. To support automation of policy decisions, it becomes crucial to encode and leverage the polygons for these relevant locations.

The Census Bureau is the United States agency that serves as the nation's leading provider of quality data about its people and economy. Yearly, the agency publishes updated and authoritative geospatial datasets to provide meaning and context to statistical data the bureau produces. Largely published in the \mytexttt{shapefile} format, the published data\footnote{\url{https://www.census.gov/programs-surveys/geography/geographies/mapping-files.html}} does include State boundaries and military installations, conveniently supporting our policy use-case. The State dataset is composed of 56 polygons, representing the 50 U.S. States, District of Columbia, plus 5 U.S. territories. The military installation dataset has 859 polygons, describing information about airports, laboratories, training areas, etc. In addition to the polygons, the datasets contain some minimal metadata about the locations, including a unique ID, and a legal name.

We have applied the \textit{Semantic Extract, Transform, and Load-r} (SETLr~\cite{mccusker_setlr_2018}) to these datasets. SETLr orchestrates ETL pipelines by the use of a script in Turtle format that defines data sources, extract and transform processes, and destination formats. SETLr was executed in both geospatial datasets to extract  shapes' information and transform them into geographical features using the PROV-O~\cite{noauthor_prov-o_nodate} (\mytexttt{prov:Location}) and GeoSPARQL (\mytexttt{geo:Feature}, \mytexttt{sf:Geometry}) ontologies. Because each phase of the ETL pipeline in SETLr is defined as an RDF resource, the complete provenance of how these geographical features came to be is maintained, thereby supporting the explanation of policy decisions in more complex scenarios where multiple locations sources are involved.

\section{Geospatial Reasoning on Radio Spectrum Policies}

Geospatial reasoning is a crucial capability when evaluating policies. Many policies, including those that regulate radio spectrum usage, are only applicable when their specified location rules are satisfied. These locations include named locations that can be mapped to features from geospatial datasets, and polygons defined directly in the policy's rules. Either way, location rules need to be correctly evaluated, taking into consideration which polygons the policy regulates, as well as coordinates that are subject to evaluation (e.g. where a radio transmission is to occur).

\begin{figure}
\vspace{-0.15in}
\centering
\includegraphics[width=0.75\textwidth]{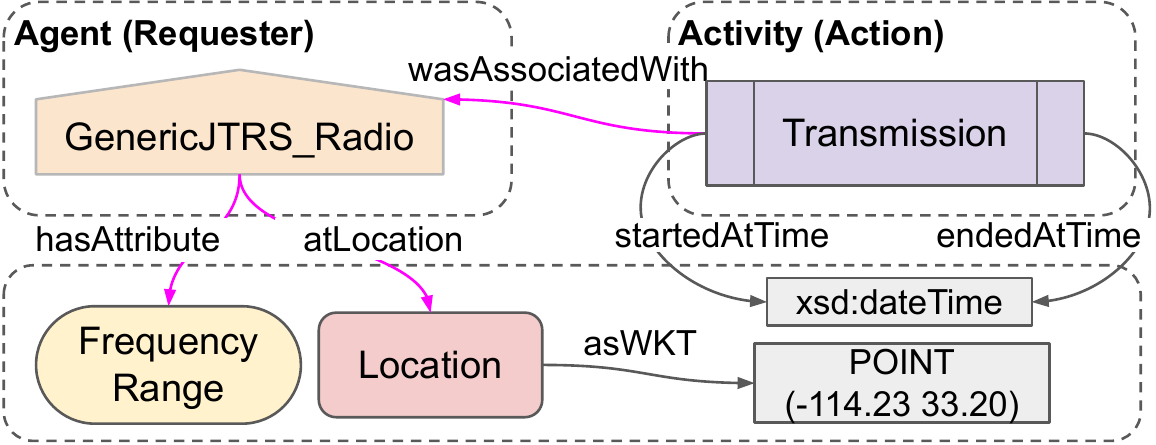}
\vspace{-0.15in}
\caption{The DSA request model}
\vspace{-0.15in}
\label{fig:request}
\end{figure}

We designed the DSA Policy Framework~\cite{santos_dsa_2020} to serve as a machine-readable, radio spectrum policy repository that can be used to automatically process radio transmission requests. The framework utilizes the World Wide Web Consortium's (W3C) OWL 2 and PROV-O, and the Open Geospatial Consortium's (OGC) GeoSPARQL 1.0 standards as a modeling foundation of radio spectrum policies and involved entities. Figure~\ref{fig:request} shows the RDF model of a transmission request within the DSA Policy Framework. Transmission requests are defined as \mytexttt{prov:Activity}, with the associated requester as a \mytexttt{prov:Agent}. Attributes that further characterize the transmission are represented using either PROV-O (including the location attribute) or a domain ontology. Coordinates in which requesters are located are represented as \textit{Well-Known Text} (WKT)~\cite{noauthor_isoiec_nodate} string, and expressed using the \mytexttt{geo:asWKT} predicate.

To allow the evaluation of the relationships between coordinates from transmissions and policy-regulated locations, we have pursued the representation of these locations as an OWL ontology where classes represent policy locations. To exemplify this approach, we will use the second provision of the US91 policy from the NTIA Redbook, which reads (with adaptations):

\begin{quote}
    \textit{``In the sub-band 1761-1780 MHz, Federal earth stations in the space operation service may transmit at the following 25 sites and non-Federal base stations must accept harmful interference caused by the operation of these earth stations: Fairbanks, Camp Parks, ... .''}
\end{quote}

Besides the policy text itself, which explicitly lists 25 sites where the policy is applicable, US91 is listed in the NTIA Redbook under the \textit{United States table}, because it is only applicable in the US and not internationally. Listing~\ref{lst-location} shows the representation of the involved locations for supporting this policy. Lines 1-7 define the \mytexttt{USLocation} class for expressing the entire United States land. This class is defined as a \mytexttt{prov:Location} and is a union of all States, District of Columbia, and territories from the appropriate Census.gov dataset, using the \mytexttt{geo:sfWithin} predicate from GeoSPARQL. Similarly, lines 9-16 extend this class to express the specific locations the above policy regulates, this time using features from the military facilities Census.gov dataset.

\lstset{language=Manchester, basicstyle=\ttfamily\fontsize{9}{10}\selectfont, columns=fullflexible, xleftmargin=5mm, framexleftmargin=5mm, numbers=left, stepnumber=1, breaklines=true, breakatwhitespace=false, numberstyle=\ttfamily\fontsize{9}{10}\selectfont, numbersep=5pt, tabsize=2, frame=lines, captionpos=b, caption={OWL expression of part of the US91 policy in Manchester syntax}, label=lst-location}
\begin{lstlisting}
Class: USLocation
  EquivalentTo:
    prov:Location and (geo:sfWithin STATE_01 or
                             geo:sfWithin STATE_02 or
                             ...
  SubClassOf:
    prov:Location

Class: US91-2-c_Location
  EquivalentTo:
    USLocation and (
    (geo:sfWithin value Fairbanks) or
    (geo:sfWithin value CampParks) or
    ...
  SubClassOf:
    USLocation
\end{lstlisting}
\vspace{-0.15in}

\subsection{Evaluating geospatial rules in policies}

To evaluate coordinates in transmission requests with policy-regulated locations, we used the GeoSPARQL function predicates embedded in SPARQL queries, as seen in Listing~\ref{lst-sparql}. The implemented queries focus on the \mytexttt{within} and \mytexttt{distance} relationships. The queries infer triples in the format \mytexttt{:req\_location geo:sfWithin :NAMED\_LOCATION}, or as a distance attribute with the numerical distance as a value and in relation to some named location.

\lstset{language=sql, basicstyle=\ttfamily\fontsize{9}{10}\selectfont, columns=fullflexible, xleftmargin=5mm, framexleftmargin=5mm, numbers=left, stepnumber=1, breaklines=true, breakatwhitespace=false, numberstyle=\ttfamily\fontsize{9}{10}\selectfont, numbersep=5pt, tabsize=2, frame=lines, captionpos=b, caption={GeoSPARQL statements to elicit select geospatial relationships.}, label=lst-sparql}
\begin{lstlisting}
FILTER(geof:sfWithin({{WKT_STR}}^^geo:wktLiteral, ?wkt))
BIND(geof:distance({{WKT_STR}}^^geo:wktLiteral, ?wkt,
  units:kilometer) AS ?distance)
\end{lstlisting}

The inferred triples are asserted back into the transmission request RDF model, which then gets reasoned over by an OWL reasoner. Using those inferred assertions, the location specified in the request can now be correctly reasoned to belong to one or more location classes, such as those in Listing~\ref{lst-location}. To exemplify this process, the coordinates for the request in Figure~\ref{fig:request} are located in Arizona. Because US91's second provision does not include any Arizona locations, no triple linking the request location to one of the policy's locations would be inferred. But, a triple linking the request location to the State of Arizona would exist (\mytexttt{:req\_location geo:sfWithin :STATE\_04}). In this setting, the request location would be reasoned to belong to the \mytexttt{USLocation} class, but not to the \mytexttt{US91-2-c\_Location} class, indicating that the transmission is to occur in the United States, but the second provision of US91 is not applicable.

Conversely, if the request in Figure~\ref{fig:request} is modified to a coordinate within the ``Fairbanks'' named location, a triple \mytexttt{:req\_location geo:sfWithin :Fairbanks} will exist. Therefore, the request location will be reasoned to belong to the \mytexttt{US91-2-c\_Location} class, making the second provision of US91 applicable.

\section{Evaluation}

We worked with radio spectrum domain experts to elicit a set of geographical requirements that a machine-readable policy model needs to support. They appear in bold in the first column of Table~\ref{tab:evaluation}. The table contains columns for Policy Representation and Request Evaluation. ``Yes'' indicates that the policy construct is either \emph{Relevant} or it has been fully addressed and \emph{Implemented}. ``Partial'' indicates that the current implementation meets a simplified requirement.

\begin{table}[ht!]
\vspace{-0.15in}
  \centerfloat
  \resizebox{.9\textwidth}{!}{
  \begin{tabular}{l|c|c|c|l}
    \toprule
    & \multicolumn{2}{c}{\textbf{Policy Representation}} & \multicolumn{2}{c}{\textbf{Request Evaluation}} 
    \\\hline
    \textbf{Locations} & \textbf{Relevant} & \textbf{Implemented} & \textbf{Relevant} & \textbf{Implemented} \\ \hline
    \ \ Named locations     & yes  & yes     & yes & yes        \\\hline
    \ \ Relative locations  & yes  & partial & yes & partial    \\\hline
    \ \ Polygons/Circles    & yes  & yes     & yes & yes        \\\hline
    \textbf{Geographical rules} \\ \hline
    \ \ Specific location   & yes  & yes     & yes & yes        \\\hline
    \ \ List of locations   & yes  & yes     & yes & yes        \\
    \bottomrule
\end{tabular}}
\caption{Geospatial semantics coverage}
\label{tab:evaluation}
\vspace{-0.25in}
\end{table}

Most policies refer to locations by names or by coordinates (points, polygons, and circles), but sometimes a location is expressed in relation to another location. Currently, relative locations have been constrained to the ones expressed using the distance relationship. Geographical rules are defined in terms of the requester being in a location or a list of locations. Our approach implements these constructs using the \mytexttt{geo:sfWithin} predicate and OWL unions.

\section{Related Work}

The works in \cite{kyzirakos_geotriples_2018,conf/edbt/PatroumpasAGA14} proposed approaches for converting geospatial content to RDF, using mapping languages and ETL pipelines. The work in \cite{bereta_ontop-spatial_2019} allows the access of geospatial datasets, including \mytexttt{shapefiles}, using an ontology-based data access approach. Our conversion relied on SETLr, which enables the data conversion of geospatial data to RDF similar to the first two approaches, but also allows the maintenance of data transformation provenance. This maintenance is important in this use-case for supporting the explanation of policy decisions.

XACML 3.0, the eXtensible Access Control Markup Language~\cite{noauthor_extensible_nodate}, is a well-known policy language and \textit{de facto} standard for representing attribute-based access control (ABAC)~\cite{hu_abac_2015} policies and requests. Importantly, XACML provides a reference architecture for centralizing access control and a process model for evaluating requests against existing policies that inform the design of access control systems across domains and technologies. Thi~\cite{tran_thi_x-strowl_2012} proposes an OWL-based extension to XACML to support a generalized, context-aware, role-based access control (RBAC) model, providing Spatio-temporal restrictions and conforming with the NIST RBAC standard~\cite{ferraiolo_rbac_2009}. Their work augments the XACML architecture with new functions and data types.

Our approach combines OWL, PROV-O, and GeoSPARQL to encode geospatial features, and an OWL reasoner to realize location class memberships. Our representation builds on previous work by matching the cross-domain policy expression semantics of XACML, extending it with the capacity to express rich Spatio-temporal restrictions, enabling the implementation of a wide variety of attribute-based policies across domains.

\section{Conclusion}

This paper presents an approach for leveraging geographical features, originally in \mytexttt{shapefiles}, to support policy decisions. In the radio spectrum domain, it is commonplace for policies to regulate the usage of the spectrum in specific locations, therefore requiring spatial reasoning to identify relationships between radio transmitters' coordinates and policy-regulated regions. This approach is an integral part of the DSA Policy Framework, which is functioning as a prototype policy management system in support of spectrum sharing operations.

Future work involves the research and development of the application of more spatial relationships, including relative locations. Besides, in other policy publications, we have encountered locations that are expressed in unusual shapes. These include paths, cones, and altitudes. More research is necessary to assess the impact in both modeling and reasoning, should we pursue this line of work. Finally, we are generalizing the approach to beyond radio spectrum policies by initially supporting practitioners from multiple domains in the creation of policies utilizing terminology and entities in domain knowledge graphs~\cite{falkow_adapt_2021}.

\bigskip\noindent\textbf{Acknowledgements.}
This work is partially funded through the National Spectrum Consortium (NSC) project number NSC-17-7030.

\bibliographystyle{splncs04}
\bibliography{bib}

\end{document}